\documentclass{amsart}

\usepackage{amssymb,latexsym,amsfonts,amsmath}
\usepackage{graphicx}
\usepackage{eso-pic}
\usepackage{epsfig}
\usepackage{graphicx}
\usepackage{xcolor}
\usepackage{algorithm}
\usepackage{algorithmic}
\usepackage{subfigure}

\usepackage{amssymb,latexsym,amsfonts,amsmath}
\usepackage{graphicx}
\usepackage{amsthm}
\usepackage{amsmath}

\newtheorem{assumption}{Assumption}

\newtheorem{theorem}{Theorem}[section]

\newtheorem{problem}[theorem]{Problem}

\theoremstyle{definition}

\theoremstyle{remark}

\numberwithin{equation}{section}

\begin{document}

\begin{abstract}
Autonomous systems, including robots and drones, face significant challenges when navigating through dynamic environments, particularly within urban settings where obstacles, fluctuating traffic, and pedestrian activity are constantly shifting. Although, traditional motion planning algorithms like the wavefront planner and gradient descent planner, which use potential functions, work well in static environments, they fall short in situations where the environment is continuously changing. This work proposes a dynamic, real-time path planning approach specifically designed for autonomous systems, allowing them to effectively avoid static and dynamic obstacles, thereby enhancing their overall adaptability. The approach integrates the efficiency of conventional planners with the ability to make rapid adjustments in response to moving obstacles and environmental changes. The simulation results discussed in this article demonstrate the effectiveness of the proposed method, demonstrating its suitability for robotic path planning in both known and unknown environments, including those involving mobile objects, agents, or potential threats.

\end{abstract}

\title[On-the-Go Path Planning and Repair
in Static and Dynamic Scenarios]{On-the-Go Path Planning and Repair
in Static and Dynamic Scenarios}

\author[daniel ajeleye]{Daniel Ajeleye \emph{Graduate Student Member, IEEE.}} 
\address{Department of Computer Science, University of Colorado Boulder, USA}
\email{daniel.ajeleye@colorado.edu}
\urladdr{https://www.hyconsys.com/members/dajeleye/}

\maketitle

\section{Introduction}
In robotics, a core task involves creating a trajectory that directs a robot from an initial configuration to a desired goal configuration. This essentially requires navigating paths that ensure \emph{safety}, meaning that no undesirable events occur. Previous efforts to define such paths are discussed in the literature, including works such as \cite{ajeleye2020time, ajeleye2022output, petti2005safe, salamati2022safety, pairet2021online} and related references. Depending on the problem, the focus may be on finding any collision-free trajectory or one that minimizes overall costs, taking into account factors like traversal time, risk, stealth, and visibility. Whether the task involves autonomously driving a vehicle or performing household chores such as cleaning the room, robots must navigate securely in the presence of dynamic elements such as people, pets, and vehicles. To achieve this, robots need to anticipate the movements of these dynamic obstacles and plan concise paths that avoid causing harm or inconvenience. Additionally, robots must be capable of rapidly generating new plans if a dynamic obstacle's trajectory changes from what was predicted, ensuring ongoing collision avoidance. Efficient path planning is crucial to equip robots with the agility to respond promptly and effectively to a continuously evolving environment.

Numerous planners have been developed to address the complexities of static environments, each offering distinct advantages and applications. Among these, the Dijkstra algorithm \cite{dijkstra2022note} is notable for its widespread use in finding the shortest path between two points in a graph. Its simplicity and optimality make it applicable to a variety of scenarios. However, its computational requirements can become impractical in larger environments, leading to the need for more efficient alternatives. The A* algorithm \cite{hart1968formal} presents a strong alternative, combining elements of Dijkstra's approach with heuristic techniques to significantly improve computational efficiency while maintaining optimality. Particularly well-suited for robotic systems operating within constrained static environments, A* offers a valuable solution when computational resources are a key consideration.

In addition to the widely used Dijkstra and A* algorithms, a range of motion planning strategies has been developed to address various challenges and specific applications. For instance, the wavefront algorithm \cite{khatib1986real} employs a distinctive method by propagating wavefronts outward from the starting point, creating a gradient map that helps in efficiently identifying the shortest path. On the other hand, the Potential Gradient Descent algorithm \cite{cauchy1847methode}, inspired by physical principles, treats the environment as a potential field. This approach guides the robot along the steepest gradient towards its goal, effectively enabling smooth and continuous navigation across different terrains \cite{kavraki1996probabilistic}. These diverse motion planning algorithms highlight the adaptability and breadth of approaches available in the field, each tailored to meet specific requirements and scenarios, thereby enriching the toolkit for robotic systems.

Additionally, discrete search algorithms operate based on the discretization of the state space traversed by the robot, utilizing the concept of finite abstraction \cite{ajeleye2023data, ajeleye2024cdata}. Examples of these algorithms include Rapidly Exploring Random Trees (RRT) \cite{lavalle1998rapidly} and its variants, such as RRT* \cite{karaman2011sampling} and RRT-Connect \cite{kuffner2000rrt}, which have become widely adopted in robotics, especially for navigating high-dimensional configuration spaces. RRT-based methods are particularly useful in situations where the precise geometry of the environment is unknown, enabling dynamic exploration and discovery of feasible paths. Similarly, Probabilistic Roadmap (PRM) algorithms \cite{kavraki1996probabilistic} are essential for constructing a detailed roadmap of the environment, facilitating efficient path planning through interconnected nodes that represent feasible configurations. PRM algorithms are notably effective in high-dimensional spaces and exhibit versatility across various robotic platforms. The selection of a specific planner depends on the task requirements, environmental characteristics, and available computational resources. Ongoing research in this field aims to refine existing planners and develop new strategies to address motion planning challenges in diverse and complex scenarios \cite{lavalle2001randomized}.

While existing planners excel in static environments, the need for dynamic planning becomes clear due to the inherently changing nature of real-world settings. The appearance of sudden obstacles or changes in terrain highlights the necessity for adaptive planning strategies. Traditional planners often struggle to handle dynamic environments effectively, underscoring the need for motion planning algorithms that can adjust paths in real-time. This adaptability is crucial for autonomous systems navigating unpredictable and constantly evolving surroundings, ensuring both safety and efficiency in the face of unforeseen obstacles. As technology advances and autonomous systems become more prevalent, the demand for robust dynamic planning algorithms is growing. This underscores the importance of continued research and development to address the challenges posed by dynamic and unpredictable real-world scenarios.

Moreover, in this work, the primary objective is to enhance traditional motion planning algorithms by integrating a mechanism designed to anticipate and adapt to the movements of dynamic obstacles. The core of the approach proposed in this work is based on the premise that environmental changes exhibit periodicity within a defined time frame during motion. As a result, the proposed planner is carefully designed to engage in periodic planning and re-planning cycles across the entire planning space using a deterministic method. This systematic approach enables the algorithm to effectively generate and adjust solutions in real-time. The integration of predictive elements into the motion planning process is central to the methodology of this work, aiming to improve the algorithm's adaptability. This enhancement ensures the algorithm's resilience and responsiveness in the face of evolving scenarios with dynamic obstacles, ultimately contributing to the development of more reliable and efficient autonomous navigation systems for real-world environments.

\subsection{Notation}
\label{sec2_1} 
Symbols $\mathbb{R}$, $\mathbb{R}_{>0}$ and $\mathbb{R}_{\ge0}$ represent the sets of real, positive real and non-negative real numbers, respectively. The notations $\cup$, $\cap$, and $\setminus$ indicate set union, intersection, and set difference, respectively. The symbol $\mathbb{N}$ and $\mathbb{N}_{\ge0}$, respectively, denotes the set of natural numbers excluding zero and natural numbers including zero. For $a, b \in \mathbb{N}_{\ge 0}$ with $a < b$, we use the notations $[a; b]$, $(a; b)$, $[a; b)$, and $(a; b]$ to represent, respectively, the closed, open, half-open from the right, and half-open from the left intervals in $\mathbb{N}_{\ge 0}$. Alternatively, for $a, b \in \mathbb{R}$ with $a < b$, we use $[a, b]$, $(a, b)$, $[a, b)$, and $(a, b]$ to denote the corresponding intervals in $\mathbb{R}$. For any non-empty set $Q$ and $n \in \mathbb{N}$, $Q^n$ indicates the Cartesian product of $n$ duplicates of $Q$. We use the operator $\mathcal{R}$ over a real interval $[a, b]$ as $\mathcal{R}([a, b])$ to generate uniformly a random number from the interval. Given $N$ vectors $x_i \in \mathbb{R}^{n_i}$, $n_i \in \mathbb{N}$, and $i \in \{1, \ldots, N\}$, we use $x = [x_1; \ldots; x_N]$ to denote the corresponding column vector of dimension $\sum_i n_i$. For any $\bar{p}, \bar{q} \in \mathbb{R}^n$ and relational operator $\simeq~\in \{\le, <, =, >, \ge\}$, where $\bar{p} = [p_1; \dots; p_n]$ and $\bar{q} = [q_1; \dots; q_n]$, $\bar{p} \simeq \bar{q}$ is interpreted as a componentwise comparison of $p_l \simeq q_l$ for all $l \in \{1, \dots, n\}$. Assuming $\bar{p} < \bar{q}$, the \emph{compact hyper-interval} $[\bar{p}, \bar{q}]$ is given as $[p_1, q_1] \times \cdots \times [p_n, q_n]$. Furthermore, given $c = [c_1; \dots; c_n] \in \mathbb{R}^n$, we define the sum $\oplus$ as $c \oplus [\bar{p}, \bar{q}] := [p_1 + c_1, q_1 + c_1] \times \cdots \times [p_n + c_n, q_n + c_n]$. For any $\bar{r} \in \mathbb{R}^n_{>0}$ and $c_0 \in \mathbb{R}^n$, notation $\Phi_{\bar{r}}(c_0)$ is interpreted as $c_0 \oplus [-\bar{r}, \bar{r}]$. For a given compact hyper-interval $H$ and discretization parameter vector $\eta_h \in \mathbb{R}^n_{>0}$, we create a partition of $H$ into cells $\Phi_{\eta_h}(h)$ such that $H \subseteq \bigcup_{h \in [H]_{\eta_h}} \Phi_{\eta_h}(h)$, where $[H]_{\eta_h}$ represents a finite set of representative points selected from those partition cells.

\section{Motion Planning}
In this section, we explore the fundamental concepts that underpin the execution of motion planning tasks.
\subsection{Preliminaries and Definitions}
A motion planning problem fundamentally involves computing a continuous path that connects a given start configuration, denoted as $X_s$, to a target goal configuration, $X_g$. The primary challenge is to find this path while avoiding collisions with known, finite obstacles, $O_i \subseteq X_o~\text{ where } i\in[1;N]$, within a defined state space $X$, commonly referred to as the environment. The geometry of both the robot and the obstacles is described within a 2D or 3D workspace, while the path itself may reside in a higher-dimensional configuration space. In this context, a configuration precisely defines the robot's pose, and the configuration space $C$ encompasses all possible configurations of the robot.

In this configuration space, the subset of configurations that avoids collisions with obstacles is known as the free space, denoted as $C_f$. The complement of $C_f$ within $C$ is referred to as the obstacle or prohibited region. Similarly, in the state space, the free space is defined as $X_f := X \setminus \bigcup_{i\in[1;N]} O_i$. Additionally, a target space, which is a subset of the free space, represents the area where the robot is intended to navigate. This target space includes the goal configurations, offering a clear spatial representation of the objective in the motion planning task.

\subsection{Problem Definition}
The core challenge arises from the inherent limitations of conventional motion planning algorithms when dealing with dynamic environments. Although traditional planners excel in navigating static spaces, their effectiveness significantly diminishes in scenarios involving moving obstacles. This shortfall is due to their lack of adaptability, making it difficult to respond to changes that occur during the execution of a motion plan. The essence of the problem highlights the need for an innovative approach that incorporates predictive capabilities, enabling real-time adjustments and repairs. This ensures the algorithm remains responsive to the dynamic nature of the environment. Additionally, we will clarify the key assumptions that form the foundation of this work.

\begin{assumption}
    \label{asum1}
    The assumptions guiding this work are outlined as follows:\\
    \begin{itemize}
        \item The environment remains stable for a specific time span $\mathcal{T}$ before undergoing any changes in its configuration. This duration, $\mathcal{T}$, is referred to as the distortion time for the dynamic environment.
        \item The distortion an environment undergoes may involve the random inclusion of obstacles into its configuration or the addition of obstacles based on mathematical models. This includes adversarial scenarios where obstacles follow a dynamic pattern.
    \end{itemize}
\end{assumption}

Building on the outlined assumptions, the main contribution of this work lies in the development of a motion planning algorithm that surpasses the limitations of traditional methods by incorporating a predictive component to effectively manage dynamic obstacles. The proposed algorithm anticipates and adjusts to the movements of dynamic elements within the environment, enabling the planner to proactively respond to changes during motion execution. By periodically engaging in deterministic planning and re-planning across the entire configuration space, the algorithm offers a robust solution capable of generating and repairing paths in real-time. The motion planning approach employed here involves decomposing the desired movement task into discrete actions that adhere to movement constraints while potentially optimizing certain aspects of the motion. This contribution is expected to significantly improve the adaptability and efficiency of autonomous systems navigating dynamic environments. We now proceed to formalize the primary problem addressed in this work.

\begin{problem}
\label{prob1}
Consider an environment $X$ with a set of finite obstacles $O_i \text{ for } i\in[1;N]$. Suppose this environment is subject to the conditions described in Assumption \ref{asum1}. The goal is to develop a motion planning algorithm that can successfully guide a robot from an initial point $x_0 \in X_s$ to a target point $x_g \in X_g$. The algorithm must navigate through the free space $X_f$ while effectively handling the inherent dynamic changes that $X$ may experience, in accordance with the distortion time and the environmental modifications described in Assumption \ref{asum1}.
\end{problem}

\section{Environmental Setup}
In this section, we introduce four primary motivating scenarios for the environment, each consistent with Assumption \ref{asum1}, to address Problem \ref{prob1}. These scenarios are described as separate cases in the following subsections.

\subsection*{Case 1}
In this scenario, we examine a dynamic environment characterized by the random appearance of obstacles during the agent's movement. These obstacles emerge unpredictably within a designated region, introducing an element of chance. Specifically, within a known compact area, obstacles appear randomly according to a probabilistic process.

Consider an autonomous exploration scenario set in an urban environment, where a delivery robot navigates a bustling city center to complete its delivery tasks. The state space is defined by the known geographical boundaries of the city, including streets, sidewalks, and public spaces. The dynamism of this environment arises from the unpredictable appearance of temporary obstacles. Examples include street vendors setting up stalls, pedestrians unexpectedly crossing the road, or construction activities introducing barriers. These factors contribute to the dynamic nature of the environment, creating a situation where obstacles emerge randomly and probabilistically during the robot's exploration.

As the delivery robot journeys through the city, it encounters sporadic changes in the landscape, mirroring the real-world challenges of navigating a dynamic urban setting. For instance, imagine the robot smoothly traversing a busy street when, suddenly, a group of pedestrians spills onto the crosswalk. Here, the motion planner, equipped with predictive capabilities, must swiftly anticipate the presence of this dynamic obstacle and adjust the robot's path in real-time. This adaptability is essential for maintaining a safe and efficient trajectory, underscoring the importance of addressing dynamic elements in the urban environment.

\begin{figure}[ht]
    \centering
    \includegraphics[width=9.0cm]{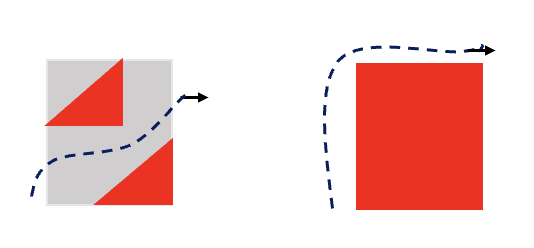}
    \caption{The static obstacle on the right necessitates the planner to circumvent with a longer path. In contrast, the left randomly appearing obstacles within a compact space, supposedly occupied by a larger obstacle, allow the planner to provide a more optimal path.
}
    \label{fc1}
\end{figure}

Notably, this scenario resembles one with static obstacles; however, the key difference is the planner's ability to utilize a potentially more optimal path, which may be shorter. This contrasts with static obstacle scenarios where planners might select less optimal routes due to the absence of real-time adjustments. To illustrate this, Fig \ref{fc1} provides a comparative example.

Furthermore, one may choose to always take the worst-case path around such a region of obstacles. However, the practical and pertinent question arises: what if such a path is unknown or difficult to find? This complication leads to the next scenario, which will be explored in Case $2$.

\subsection*{Case 2}
This encompasses an overarching scenario comprising two subcases, each of which will be explored in the subsequent sections.

\subsubsection*{Changing environment with disappearing obstacles}
In this sub-case, the known compact region undergoes dynamic transformations over time. Initially appearing obstacles emerge within the predefined spatial boundaries, adding complexity to the agent's path planning. However, as the environment changes, these obstacles vanish, creating space for a different set of obstacles to randomly appear within the same compact region. The motion planning algorithm must not only adapt to the random emergence of obstacles but also deal with the evolving nature of the environment. It should anticipate the disappearance of obstacles and efficiently recalculate paths based on the evolving configuration of the known region. This scenario adds an additional layer of complexity, demanding adaptability to both the randomness of obstacle appearance and the dynamic changes in the environment over time.

As an illustration, imagine a smart city equipped with an intelligent traffic management system. In this scenario, dynamic obstacles such as road maintenance zones or temporary road closures may emerge within a known compact region. However, these obstacles are temporary in nature and once maintenance is completed or the event causing closure is complete, the obstacles disappear. Simultaneously, the environment undergoes changes, leading to the emergence of new obstacles, such as detours for special events or spontaneous road closures for parades. This subcase encapsulates the dynamic nature of urban environments, where temporary obstacles come and go, and the motion planner must adeptly navigate these changes for efficient and reliable path planning.

\subsubsection*{Persistent obstacles with continuous emergence}
In this sub-case, the scenario unfolds in a manner in which both obstacles that appeared earlier and newly emerging obstacles coexist within the known compact region. The challenges escalate as the motion planning algorithm must navigate not only around obstacles that persist from earlier states, but also contend with the continuous influx of new obstacles. This dynamic environment demands a high level of adaptability, as the planner must factor in the presence of both existing and newly introduced obstacles while generating and adjusting paths in real-time. The coexistence of persistent and novel obstacles adds an intricate layer to the motion planning process, requiring the algorithm to make informed decisions to ensure the agent's safe and efficient navigation.

In the domain of e-commerce fulfillment centers, autonomous mobile robots traverse a known compact region defined by the layout of shelves and storage areas. Existing obstacles, such as stationary shelves or other robots, persist in their locations. Simultaneously, the continuous influx of new orders introduces dynamically emerging obstacles, manifesting themselves as temporary zones with high human activity-representing areas where workers pick items for shipping.

The motion planning algorithm in this context must appropriately account for both persistent obstacles and the continuous appearance of new dynamic obstacles. Its objective is to optimize robot routes for efficient and collision-free navigation in the evolving warehouse environment. Consequently, for every encountered blockage in the robot's path, the algorithm must facilitate a seamless rerouting of the agent to ensure uninterrupted and streamlined operations within the fulfillment center. This scenario reflects the intricacies of dynamic environments in practical settings, demanding a robust and responsive motion planning strategy for autonomous robots.

\subsection*{Case 3}
In this scenario, the environment is open, allowing obstacles to randomly emerge anywhere, presenting a dynamic and unpredictable landscape. The crucial constraint in this case is that the newly emerging obstacles focus exclusively on the portion of the environment devoid of the path covered so far by the agent. This scenario reflects a situation where the agent's planned path is considered sacred, and the motion planning algorithm must ensure the continuous avoidance of obstacles as the motion progresses. The algorithm is tasked with dynamically adapting to the emergence of obstacles in previously nontraversed areas, safeguarding the integrity of the planned path, and ensuring obstacle-free navigation as the agent moves forward.

A tangible real-world instance of this case is exemplified by autonomous aerial surveillance drones. Imagine a fleet of these drones assigned the mission of monitoring a large geographical area for security purposes. The drones diligently adhere to preplanned paths to systematically cover the entire region, with the flexibility to modify the path as needed. In this scenario, the environment is open, permitting the random emergence of obstacles, ranging from temporary structures to weather balloons or other flying objects.

The critical aspect here is that the motion planning algorithm assumes that no obstacle appears directly on the drone's pre-determined flight path. This requirement emphasizes the significance of maintaining the integrity of the planned path and covered so far, safeguarding it from any dynamically emerging obstacles. The drones must navigate the open environment with adaptability and precision, avoiding obstacles in real time while adhering to their predefined surveillance paths.

To illustrate, consider an unexpected hot air balloon ascending in the vicinity of a drone's planned trajectory. In such a scenario, the motion planning system must dynamically reconfigure the drone's path to circumvent the obstacle while ensuring the continuous coverage of the surveillance mission. This case underscores the critical need for a robust motion planning algorithm capable of not only responding to randomly emerging obstacles, but also actively preventing interference with planned paths to maintain the efficacy of the autonomous surveillance operation. With this, we transition to the final case to be considered in this work.

\subsection*{Case 4}
In this scenario, the emergence of obstacles follows a dynamic and potentially adversarial pattern, coupled with the constraint of Case 3. The motion planning algorithm assumes the role of the second player in a game-like environment, strategically responding to the evolving moves of an adversarial obstacle dynamics. The objective is to find safe paths for the autonomous agent despite the intentional efforts of adversarial obstacle dynamics to impede progress.

A tangible real-world example of this case is illustrated by autonomous vehicle security convoys operating in urban settings. Imagine a convoy of autonomous security vehicles assigned to patrol urban areas. The environment is subject to dynamic changes, and potential threats—represented by dynamically adversarial obstacle dynamics—may intentionally attempt to disrupt the convoy's path. These threats could manifest as simulated adversarial vehicles or unpredictable civilian movements, simulating a game-like scenario.

The motion planning algorithm, acting as the second player in this dynamic game, continuously assesses the evolving obstacle dynamics and strategically plans alternative routes to ensure the security convoy's safe navigation. For instance, if an adversarial vehicle attempts to block the planned route, the motion planning algorithm swiftly adapts, finding alternative paths to avoid interference and maintain the security mission's integrity. This scenario encapsulates the intricate interplay between autonomous agents and potential threats in security-related applications, where the motion planning algorithm must act dynamically and strategically to outmaneuver adversarial obstacle dynamics.

Having outlined the assumptions considered in this work along with various cases of dynamic environments, we proceed to present the motion planning algorithm in the next section.

\section{On-the-Go Motion Planning Algorithm}
In this pivotal section, we unveil the core of this work: an innovative and adaptive motion planning algorithm named the ``On-the-Go Motion Planning Algorithm.'' This approach is designed to address the challenges posed by dynamic and evolving environments, where obstacles may emerge unpredictably, necessitating instantaneous adjustments in an autonomous system's planned trajectory. The algorithm is crafted to seamlessly function within this ever-changing landscape, continually assessing and responding to the environment's dynamics.

The ``On-the-Go Motion Planning Algorithm'' introduces adaptive strategies, excelling not only in scenarios with known compact regions featuring randomly emerging obstacles but also in cases where obstacles follow adversarial dynamics. Algorithm \ref{alg1} outlines the procedures, delving into the intricacies of the motion planning strategy. It details the components, decision-making processes, and underlying principles that empower the algorithm to provide efficient, real-time solutions for autonomous navigation in dynamic environments, encompassing the highlighted cases.

   \begin{algorithm}[ht!]
	\caption{Dynamic motion planning algorithm with adaptive strategies for evolving environments\\ \texttt{// Inputs are initial point $x_0\in X_s$ of the robot, the destination point $x_g\in X_g$, case $\mathcal{K}$ describing the dynamical environment $\mathcal{E}$ scenario, a traditional planner $\mathcal{P}$, which is preferred as suitable for static environment, distortion time $\mathcal{T}$ of $\mathcal{E}$.}}
 \label{alg1}
 \begin{algorithmic}[1]
\REQUIRE \emph{Planner}\,(\,$x_0$, $x_g$, $\mathcal{E}$, $\mathcal{K}$, $\mathcal{P}$, $\mathcal{T}$) 
    \STATE \texttt{// break down the considerations in the planner to the cases}
    \STATE $i\leftarrow 1$ \texttt{// initialize a counter $i$ to track the motion steps}
        \STATE $x_c\leftarrow x_0$ \texttt{// initialize a state $x_c$ to track the current position of the agent}
        \STATE $path\leftarrow[x_0]$ \texttt{// initialize a static stationary path}
    \IF{$\mathcal{K}==Case~1$ or $2$}
        \WHILE{ $x_c$ is not yet at (approximately close to) $x_g$ }
            \STATE Plan by utilizing $\mathcal{P}$, and extend $path$ accordingly to set out for motion
            \STATE $x_c\leftarrow path[\mathcal{T}]$ \texttt{// $path[\mathcal{T}]$ is the agent's current state after the time span $\mathcal{T}$}
            \STATE $\mathcal{E}$ gets adjusted according to case $\mathcal{K}$
            \STATE $\mathcal{T}\leftarrow \mathcal{T}*i$ and $i\leftarrow i+1$
        \ENDWHILE
    \ENDIF
    \IF{$\mathcal{K}==Case~3$ or $4$}
        \WHILE{ $x_c$ is not yet at (approximately close to) $x_g$ }
            \STATE Plan by utilizing $\mathcal{P}$, and extend $path$ accordingly to set out for motion
            \STATE \texttt{// These cases are adversarial over $\mathcal{E}$, and $\mathcal{P}$ may end up loosing the game, \emph{i.e.,} finds no further path to proceed the motion}  
            \IF{$\mathcal{P}$ finds no further possible path }
            \STATE \textbf{Return:} \texttt{`Road blocked completely'}
            \ENDIF
            \STATE $x_c\leftarrow path[\mathcal{T}]$ \texttt{// $path[\mathcal{T}]$ is the agent's current state after the time span $\mathcal{T}$}
            \STATE $\mathcal{E}$ gets adjusted according to case $\mathcal{K}$
            \STATE $\mathcal{T}\leftarrow \mathcal{T}*i$ and $i\leftarrow i+1$
        \ENDWHILE
    \ENDIF
    \ENSURE $path$
\end{algorithmic}
\end{algorithm}

Remarkably, one could observe that the planner could be hindered from successfully navigating the agent to the goal. This is because the way the environment $\mathcal{E}$ gets modified is adversarial. One could think of the static environment traditional planner $\mathcal{P}$ inputted into Algorithm \ref{alg1} as an existential player, while the strategy in which the environment is modified is a universal player as it either randomly or dynamically sets obstacles across $\mathcal{E}$. Therefore for some (universal) emergence of obstacles, there may not exist a path to further the motion of the agent. 

Furthermore, we present an example of a traditional planner $\mathcal{P}$, upon drawing inspiration from discretized motion planning algorithm \cite{xu2012real,belta2005discrete}, which is outlined in Algorithm \ref{alg2}. This algorithm will be utilized later on to showcase an example in the next section together with the proposed planner in Algorithm \ref{alg1}. In addition, we highlight the helper functions adopted in Algorithm \ref{alg2} as follows. For any $A\subset X$, $IsObstacle(A)$ returns \texttt{true} if 
$A\cap O_i\ne\emptyset$ $\forall i\in[1;N]$. Likewise, $IsSatisfiesDynamicConstraint(x,\Phi_h(\hat x))$ is an auxiliary function, which produces a state (point) $x'\in X\cap \Phi_h(\hat x)$ along the trajectory of the agent, which is provided by the given kinodynamics of the agent when instantiated from $x$ over a time span.

\begin{algorithm}[ht!]
	\caption{A traditional motion planner for an evolved environment\\ \texttt{// Inputs are initial point $x_0\in X_s$ of the robot, the destination point $x_g\in X_g$, s indicates whether the planner considers a given kinodynamic constraint of the agent or not.}}
 \label{alg2}
 \begin{algorithmic}[1]
\REQUIRE \emph{$\mathcal{P}$}\,(\,$x_0,x_g\in X$, $h$, s)
\STATE \texttt{// discretize the state space $X$}
\STATE Construct $\hat{X}\leftarrow[X]_h$, $\hat x_0\leftarrow\hat x'\in\hat X$ and $\hat x_g\leftarrow\hat x''\in\hat X$ where $x_0\in\Phi_h(\hat x')$ and $x_g\in\Phi_h(\hat x'')$
\STATE $\hat X_o:=\{\hat x\in \hat X~|~IsObstacle(\Phi_{h}(\hat x))==\texttt{true}\}$ 
\STATE Create a labeling function $L:[X]_h\rightarrow\mathbb{R}_{\ge 0}$ such that $L(\hat x)\leftarrow 1~\forall\hat x\in \hat X_o$, $L(\hat x_g)\leftarrow 2$, and ensure $L(\hat x)\leftarrow n\in\mathbb{N}$ where $\forall \hat x_n\in \hat X:~\hat x\in\Phi_{2h}(\hat x)$, $L(\hat x_n)\leftarrow n+1$
\STATE $traj\leftarrow[x_0]$ \texttt{// initialize a trajectory for the agent}
\STATE Set $x_d\leftarrow \hat x_0$ and $x_c\leftarrow x_0$
\WHILE{ $x_c\notin \Phi_h(\hat x_g)$ }
 \STATE $x_d\leftarrow \hat x\in \Phi_{2h}(x_d)$ where $L(\hat x)= L(x_d)-1$
 \IF{$s$ indicates no kinodynamic constraint}
 \STATE $x_c\leftarrow x_d$
\STATE $traj\leftarrow [traj;x_d]$ 
 \ENDIF
 \IF{$s$ indicates kinodynamic constraint}
 \STATE $x_c\leftarrow IsSatisfiesDynamicConstraint(x_c,\Phi_h(x_d))$
\STATE $traj\leftarrow [traj;x_c]$ 
 \ENDIF
 \ENDWHILE
	\ENSURE $traj$
\end{algorithmic}
\end{algorithm}

\section{Experimental Setup and Results}
\begin{figure}[!ht]
    \includegraphics[width=9.0cm]{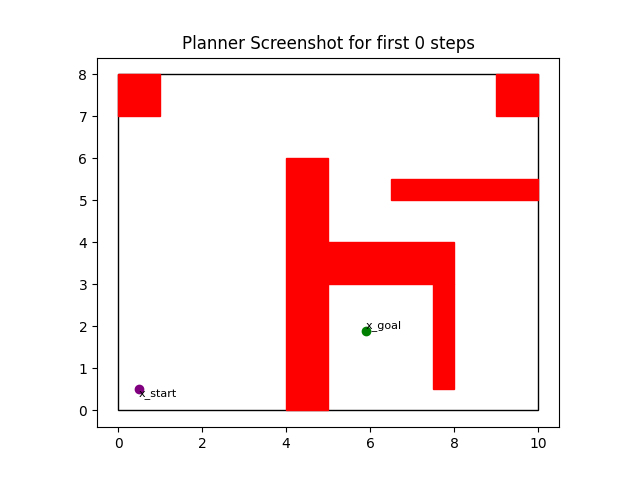}
    \includegraphics[width=10.0cm]{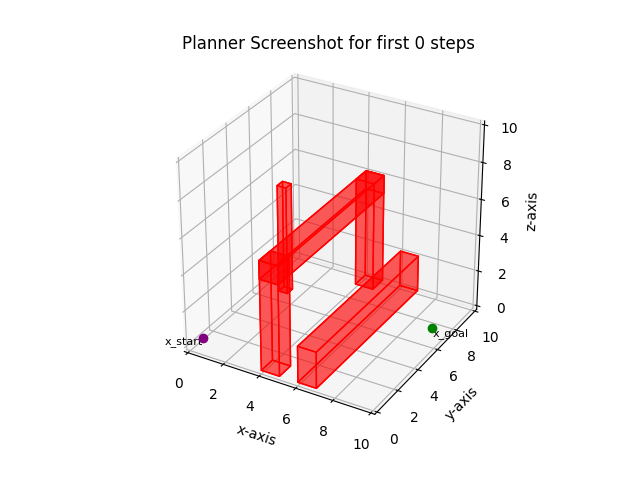}
    \includegraphics[width=10.0cm]{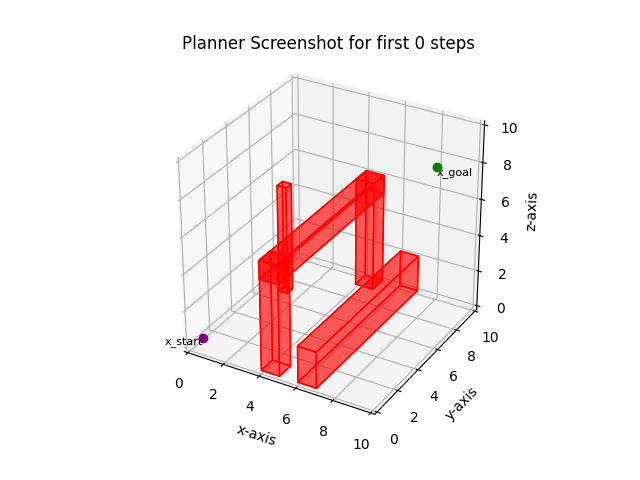}
    \caption{The top-left section illustrates the 2D environment, while the top-right and bottom sections represent the two 3D environments. In these diagrams, the purple points mark the agent's starting position, and the green points denote the target location. The red bars represent the obstacles present in each environment.}
    \label{efig1}
\end{figure}

In this section, we show the efficacy of the approach proposed in this work by applying them, based on the highlighted scenarios in the previous section, to some numerical benchmarks. All implementations in this work have been done in Python programming language, with a 64-bit MacBook Pro with $64$GB RAM ($3.2$ GHz). Furthermore, the environmental setup for the experiment is shown in Fig. \ref{efig1}. 

\begin{figure}[ht]
    \includegraphics[width=7.5cm]{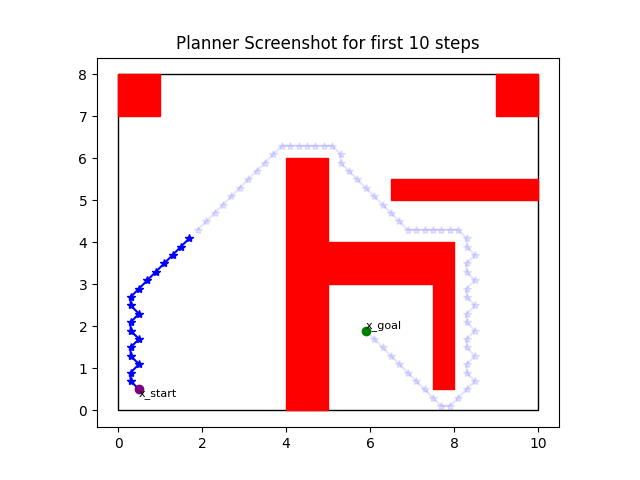}
    \includegraphics[width=7.5cm]{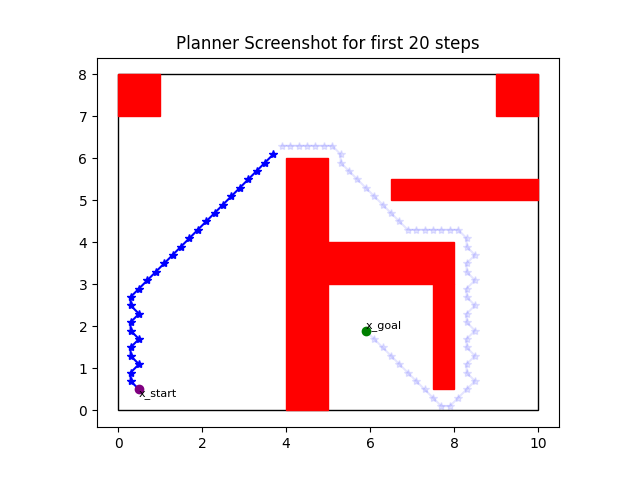}
    \includegraphics[width=7.5cm]{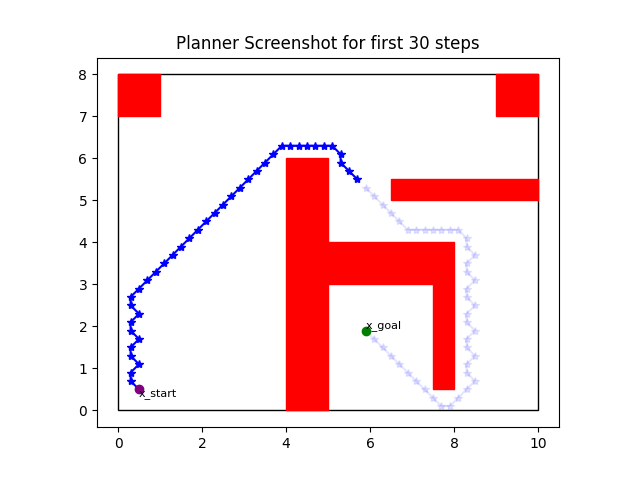}
    \includegraphics[width=7.5cm]{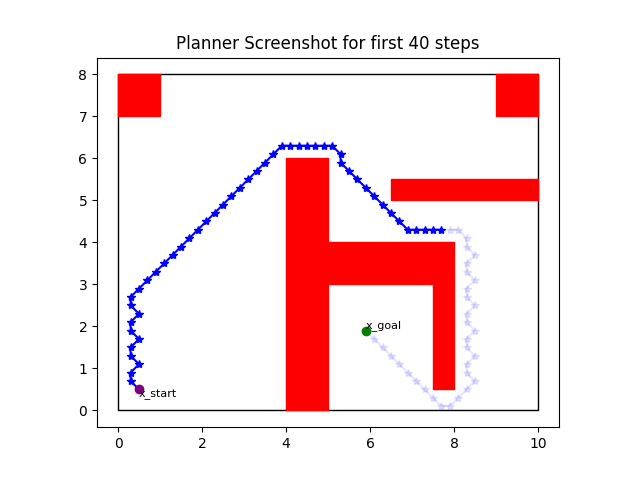}
    \includegraphics[width=7.5cm]{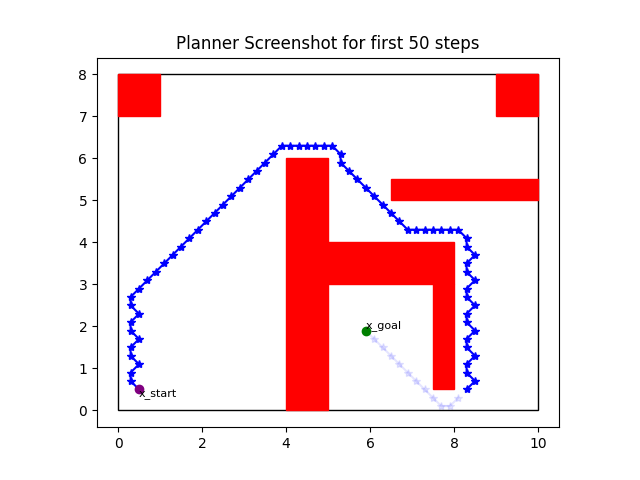}
    \includegraphics[width=7.5cm]{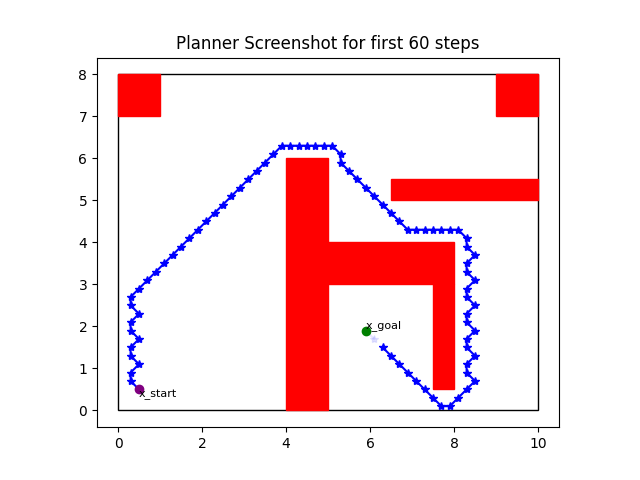}
    \includegraphics[width=7.5cm]{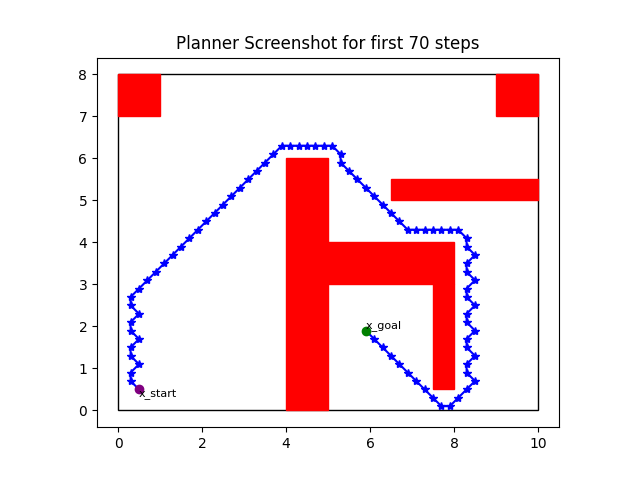}
    \caption{An illustration of the Planner on the static 2D environment as the agent progresses through successive steps until reaching $x_{goal}$. }
    \label{efig2}
\end{figure}

\begin{figure}[ht]
    \includegraphics[width=7.5cm]{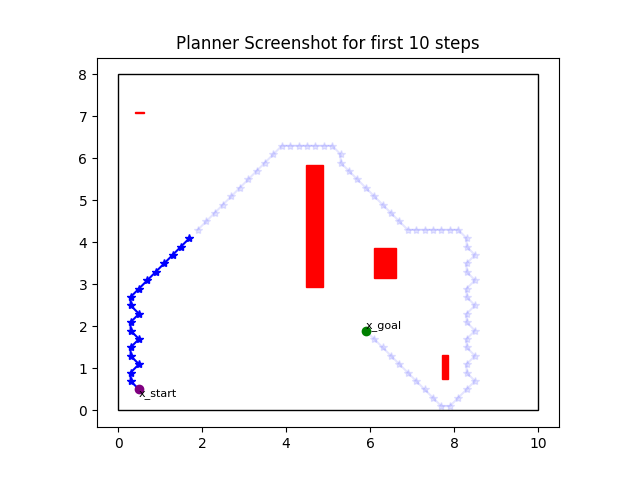}
    \includegraphics[width=7.5cm]{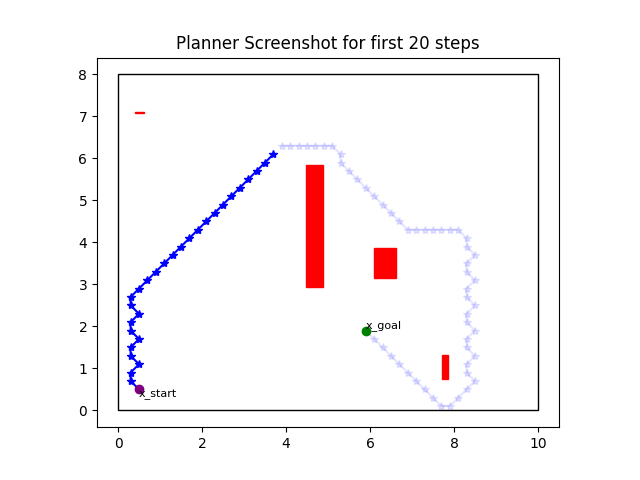}
    \includegraphics[width=7.5cm]{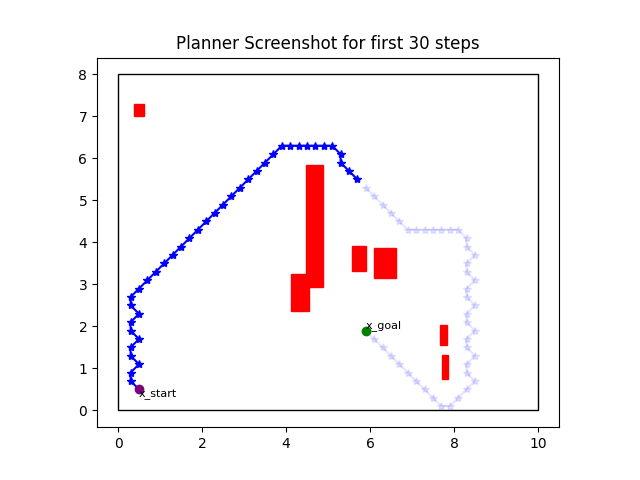}
    \includegraphics[width=7.5cm]{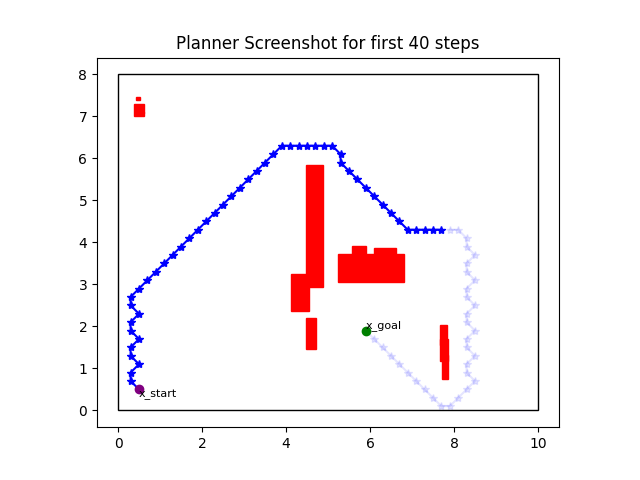}
    \includegraphics[width=7.5cm]{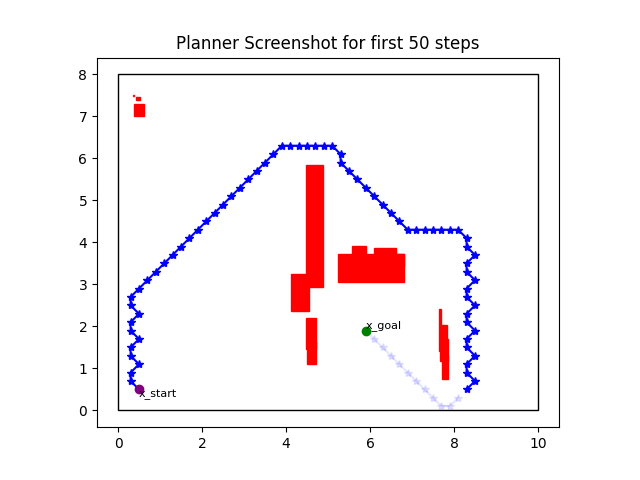}
    \includegraphics[width=7.5cm]{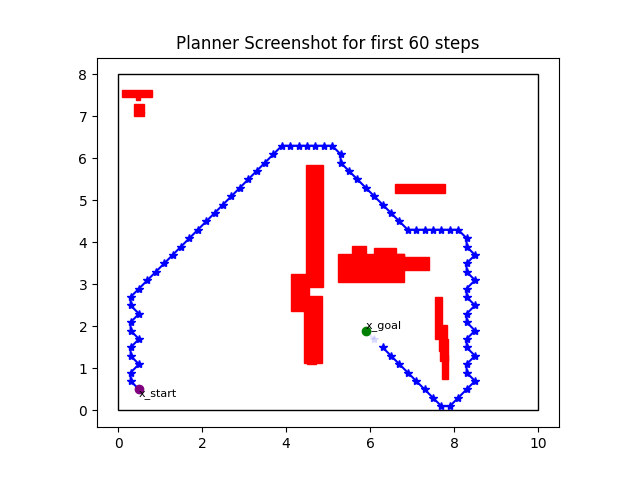}
    \includegraphics[width=7.5cm]{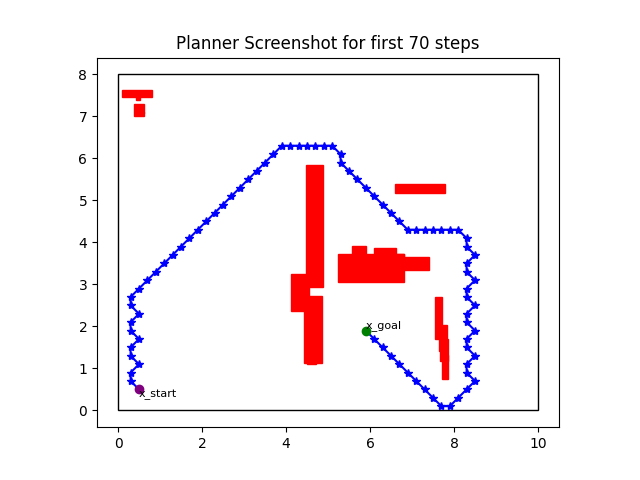}
    \caption{An illustration of the Planner on the dynamic 2D environment evolving according to Case $1$ as the agent advances through each step until reaching $x_{goal}$.}
    \label{efig3}
\end{figure}

\begin{figure}[ht]
    \includegraphics[width=7.5cm]{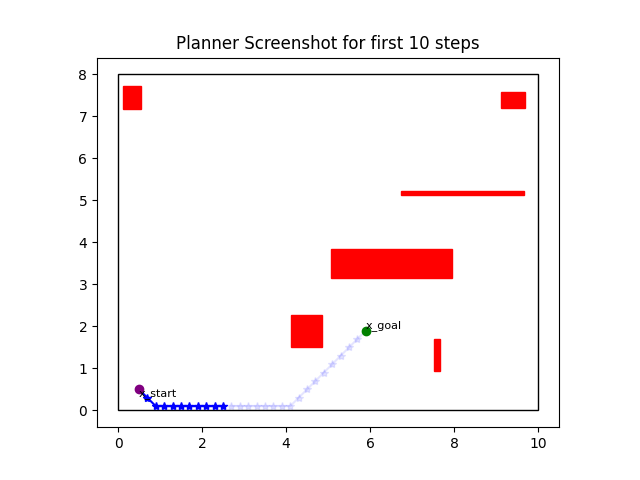}
    \includegraphics[width=7.5cm]{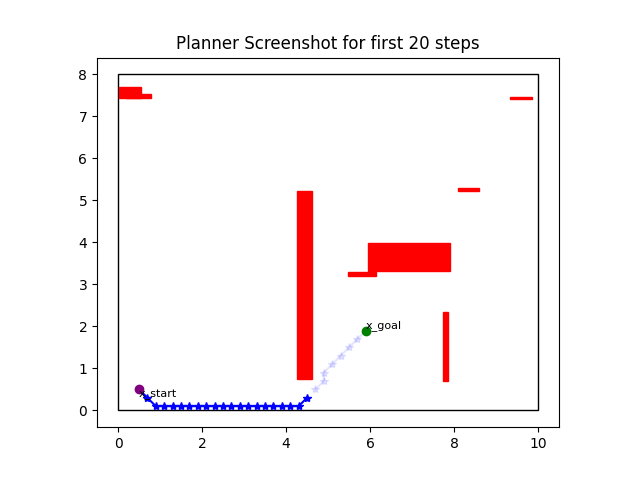}
    \includegraphics[width=7.5cm]{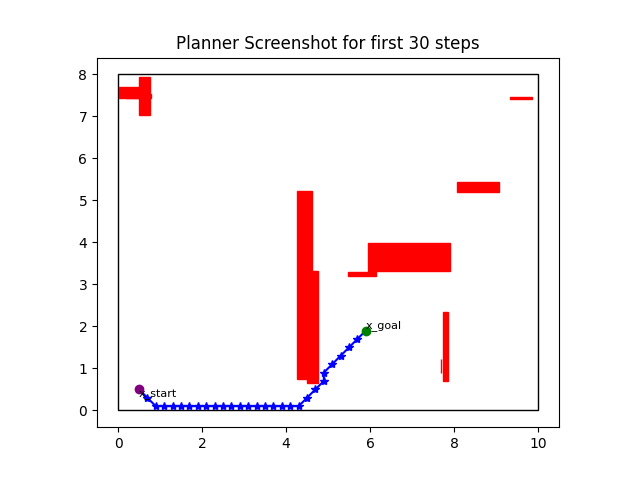}
    \caption{An illustration of the Planner on the dynamic 2D environment evolving according to the first scenario in Case $2$ as the agent progresses step by step until reaching $x_{goal}$. }
    \label{efig4}
\end{figure}

\begin{figure}[ht]
    \includegraphics[width=7.5cm]{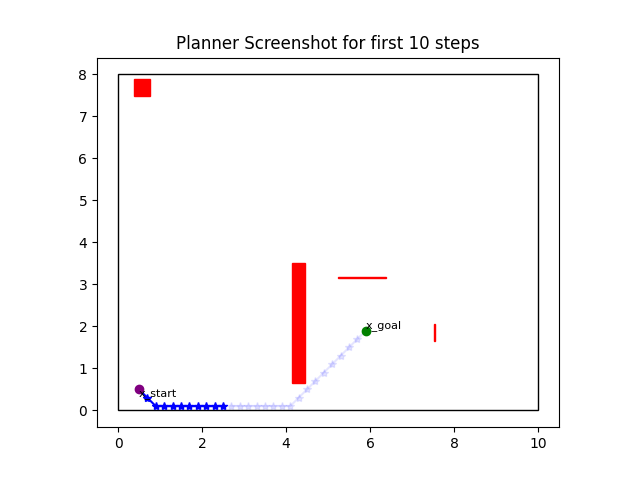}
    \includegraphics[width=7.5cm]{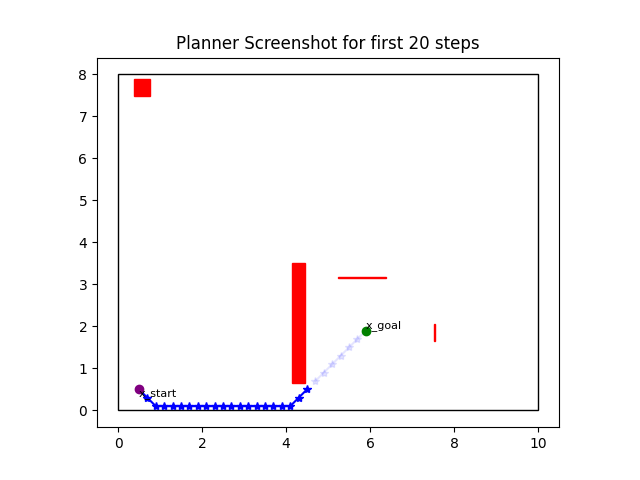}
    \includegraphics[width=7.5cm]{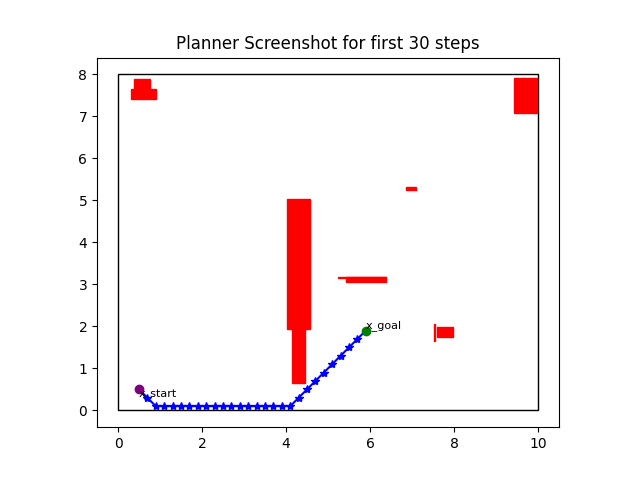}
    \caption{An illustration of the Planner in a dynamic 2D environment, evolving as described in the first scenario of Case $2$, showing the agent's step-by-step progression until it reaches $x_{goal}$.}
    \label{efig5}
\end{figure}

\begin{figure}[ht]
    \includegraphics[width=7.5cm]{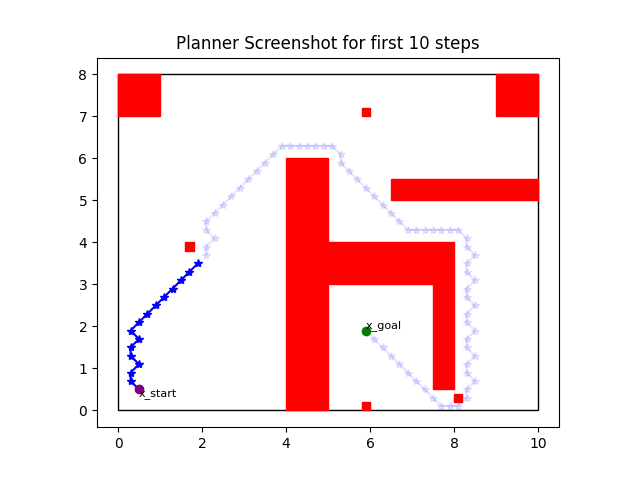}
    \includegraphics[width=7.5cm]{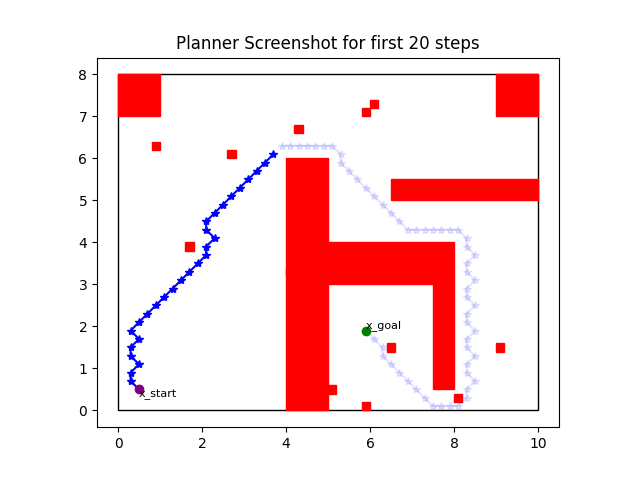}
    \includegraphics[width=7.5cm]{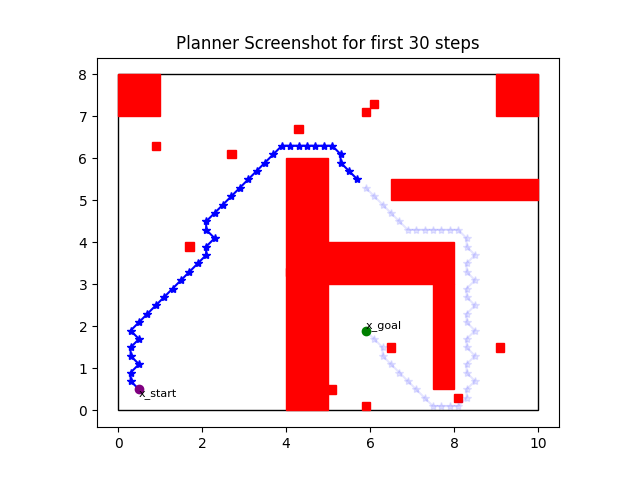}
    \includegraphics[width=7.5cm]{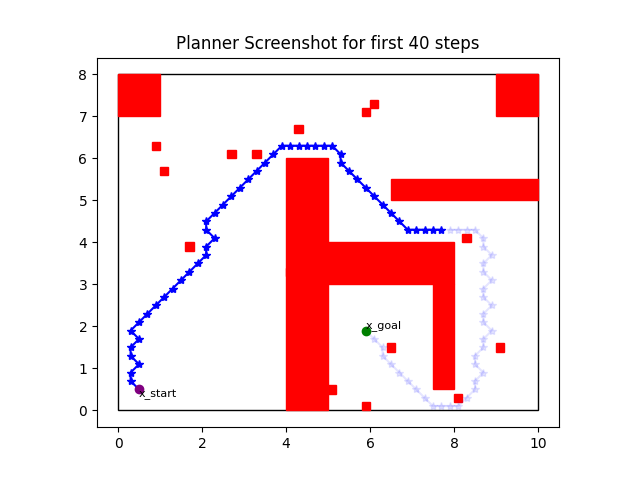}
    \includegraphics[width=7.5cm]{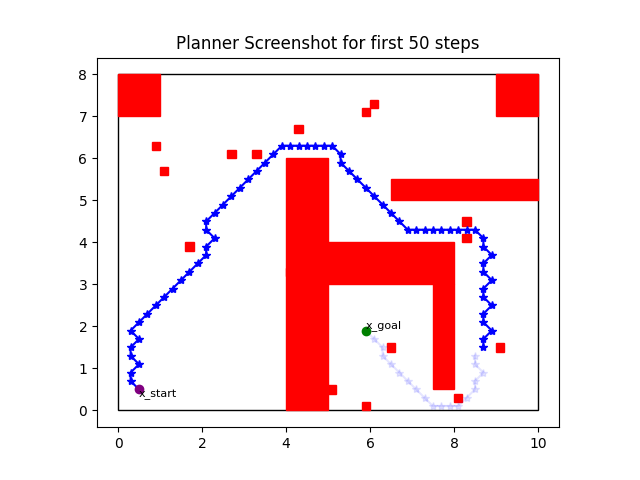}
    \includegraphics[width=7.5cm]{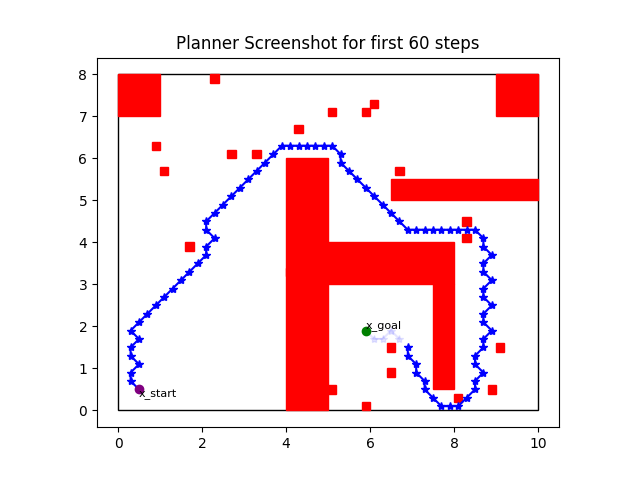}
    \includegraphics[width=7.5cm]{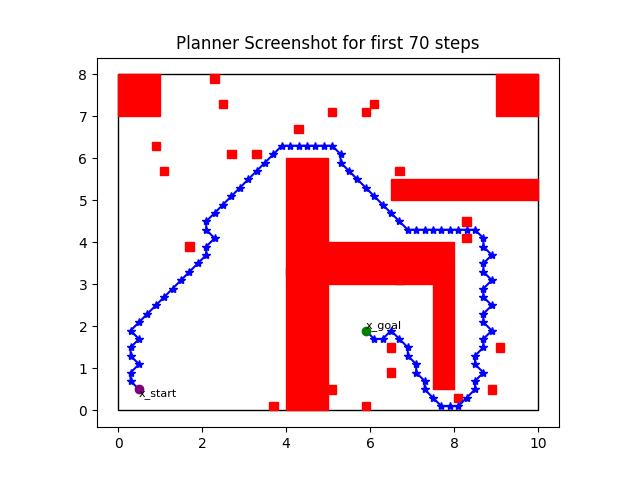}
    \caption{An illustration of the Planner on the dynamic 2D dynamic environment evolving according to Case $3$ as the agent progresses step by step until reaching $x_{goal}$. }
    \label{efig6}
\end{figure}

\begin{figure}[ht]
    \includegraphics[width=7.5cm]{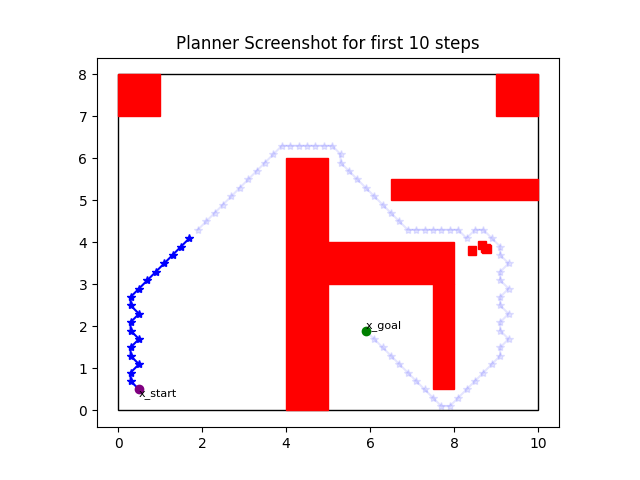}
    \includegraphics[width=7.5cm]{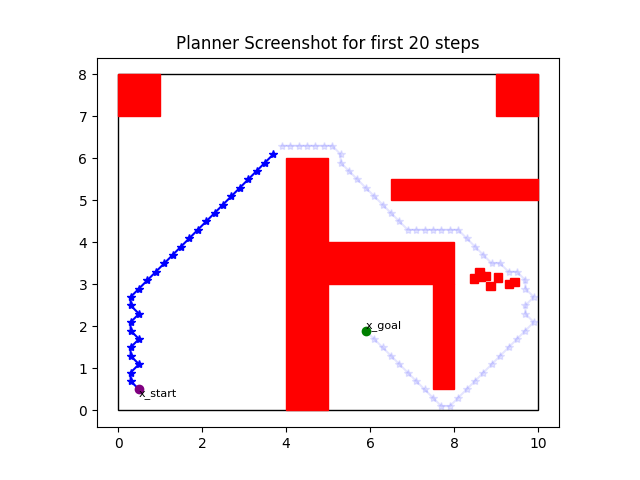}
    \includegraphics[width=7.5cm]{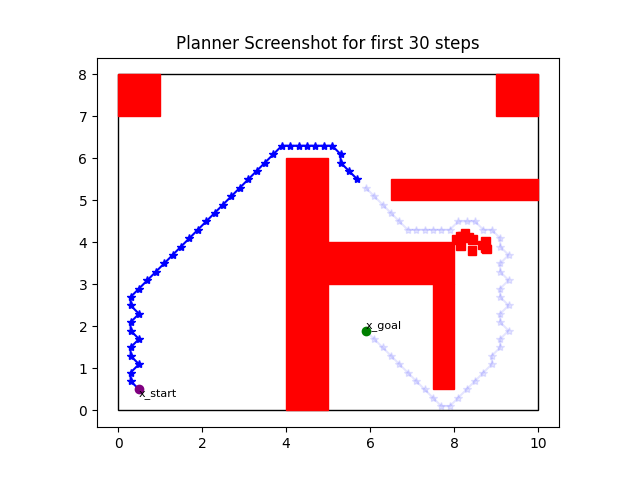}
    \includegraphics[width=7.5cm]{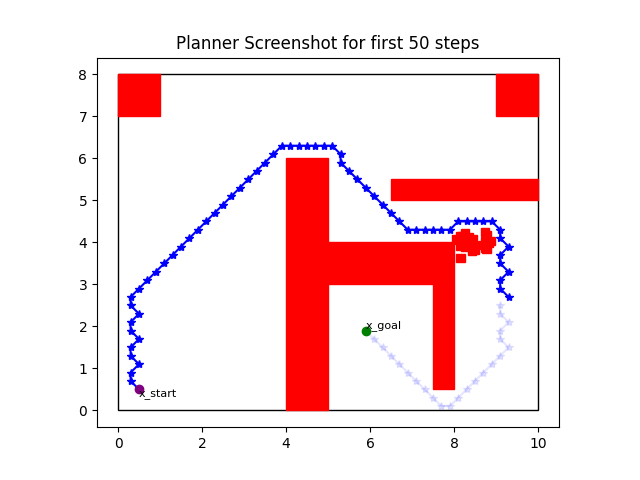}
    \includegraphics[width=7.5cm]{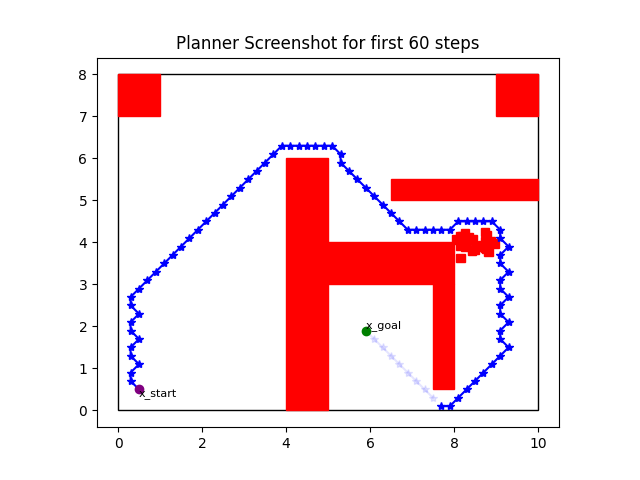}
    \includegraphics[width=7.5cm]{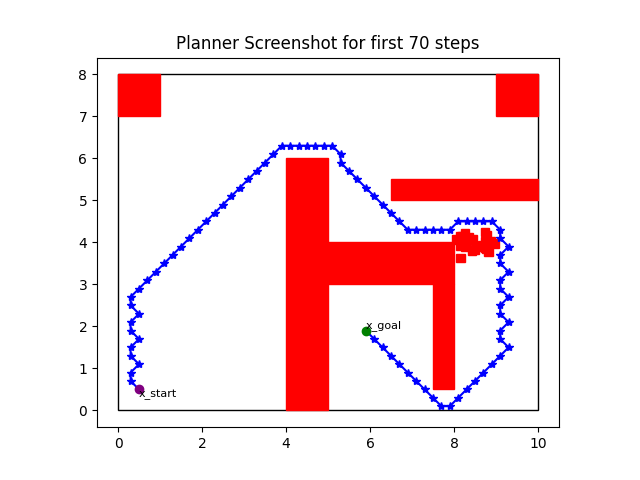}
    \caption{An illustration of the Planner on the dynamic 2D dynamic environment evolving according to Case $4$ as the agent progresses step by step until reaching $x_{goal}$. }
    \label{efig7}
\end{figure}

The cases $1$ to $4$ have been considered for the 2D example, where $X=[0,10]\times[0,8]$. The results of the simulation by Algorithm \ref{alg1} and \ref{alg2} on the 2D examples are presented in Figs. \ref{efig2}, \ref{efig3}, \ref{efig4}, \ref{efig5}, \ref{efig6} and \ref{efig7}. The `faint' blue path signifies a potential planned path, which has not been covered by the agent. We adopted the dynamics presented in \eqref{2dyn} to represent the adversarial dynamic obstacles involved used in Case $4$ for the 2D example.  
\begin{equation}
    \label{2dyn}
    \begin{cases}
        x(k+1)=x(k)+2\mathcal{R}([-h,h])\\
        y(k+1)=y(k)+\mathcal{R}([-h,h]),
    \end{cases}
\end{equation}
where $k\in\mathbb{N}_{\ge0}$ represents the time steps in motion, with state $(x,y)\in X$ and $h$ is the discretization parameter fed into the Planner, which is taken as $0.2$ for all the experiments. In addition, each of the experiments was concluded in at most $6s$, including the time to generate all the screenshots. 


\begin{figure}[ht]
    \includegraphics[width=7.5cm]{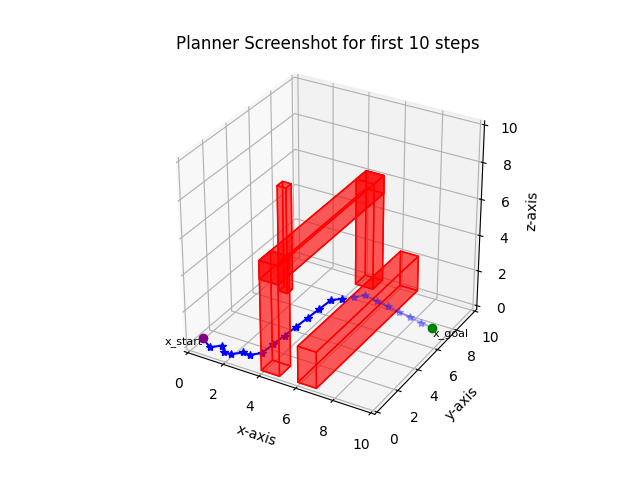}
    \includegraphics[width=7.5cm]{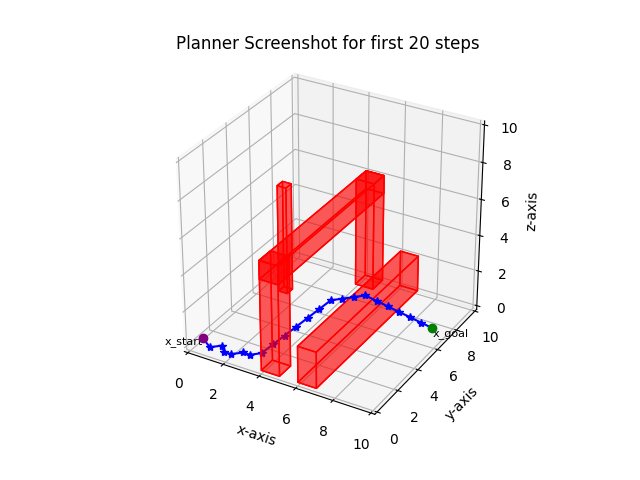}
    \caption{An illustration of the Planner, without kinodynamics constraints \eqref{drone}, on the dynamic 3D environment evolving according to Case $1$ as the agent progresses step by step until reaching $x_{goal}$.}
    \label{efig8}
\end{figure}

\begin{figure}[ht]
    \includegraphics[width=7.5cm]{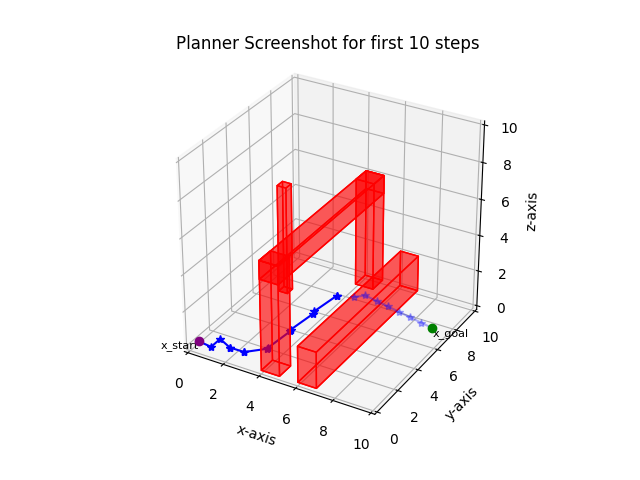}
    \includegraphics[width=7.5cm]{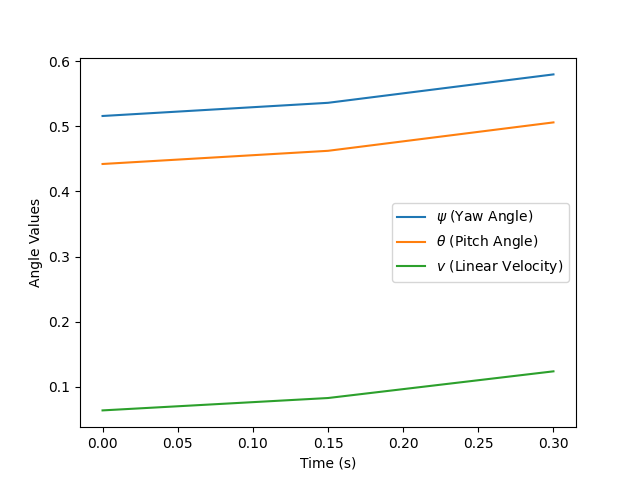}
    \includegraphics[width=7.5cm]{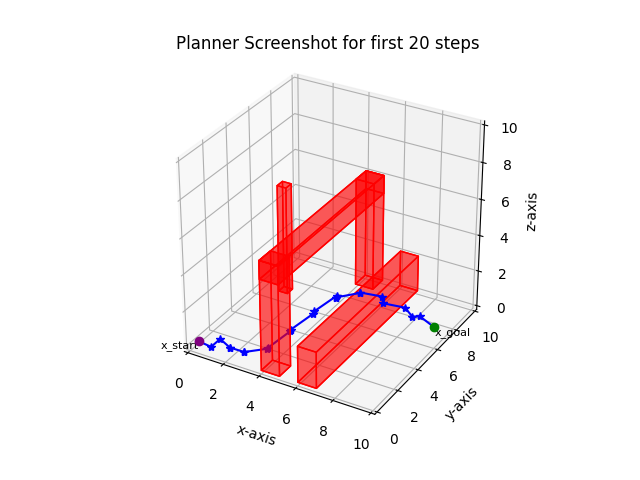}
    \includegraphics[width=7.5cm]{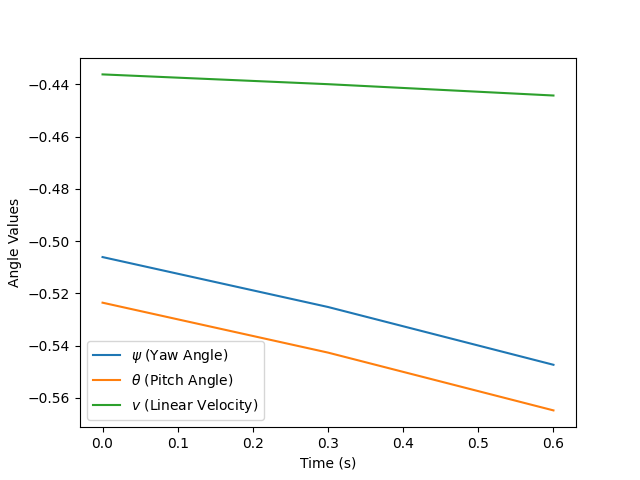}
    \caption{An illustration of the Planner, involving kinodynamics constraints \eqref{drone}, on the dynamic 3D environment evolving according to Case $1$ as the agent progresses step by step until reaching $x_{goal}$. The motion progression is displayed on the left, with the corresponding control inputs applied shown on the right.}
    \label{efig9}
\end{figure}

\begin{figure}[ht]
    \includegraphics[width=7.5cm]{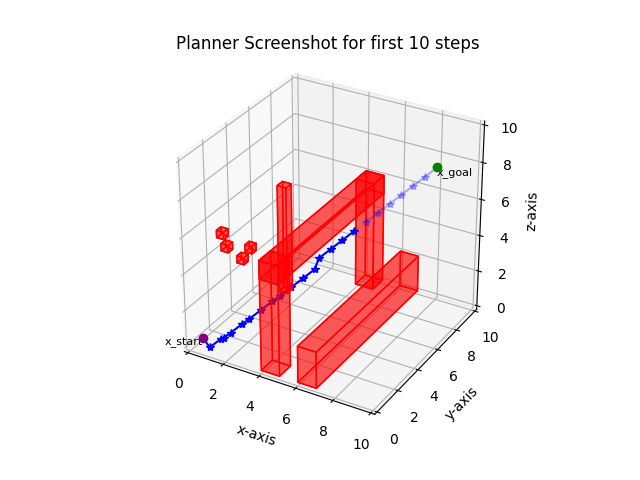}
    \includegraphics[width=7.5cm]{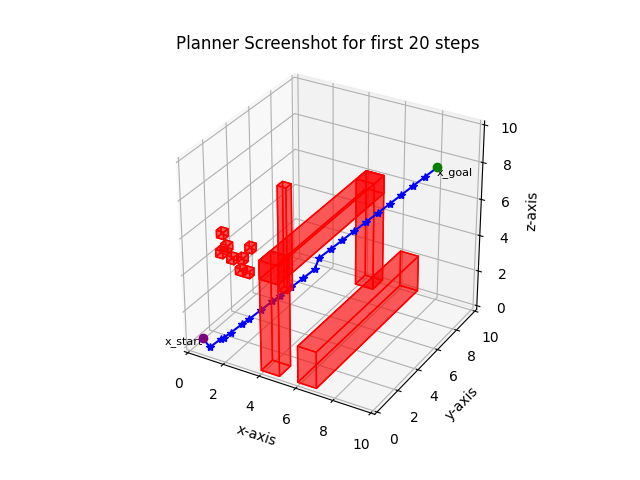}
    \caption{An illustration of the Planner, without kinodynamics constraints \eqref{drone}, on the dynamic 3D environment evolving according to Case $4$ as the agent progresses step by step until reaching $x_{goal}$.}
    \label{efig10}
\end{figure}

\begin{figure}[ht]
    \includegraphics[width=7.5cm]{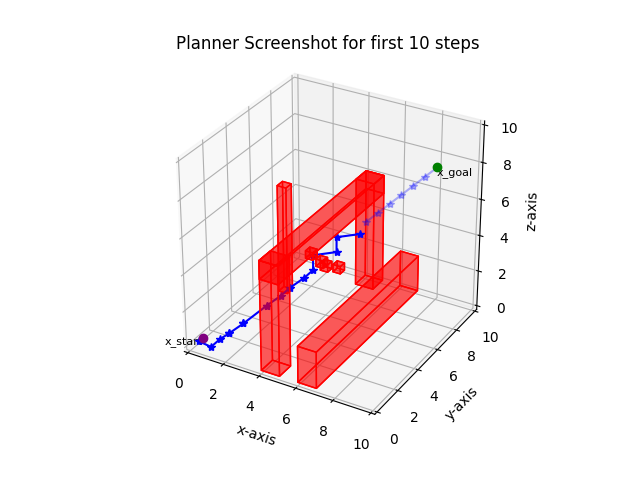}
    \includegraphics[width=7.5cm]{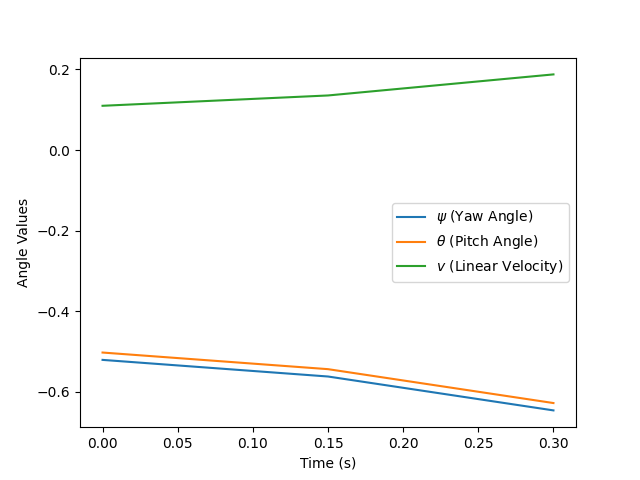}
    \includegraphics[width=7.5cm]{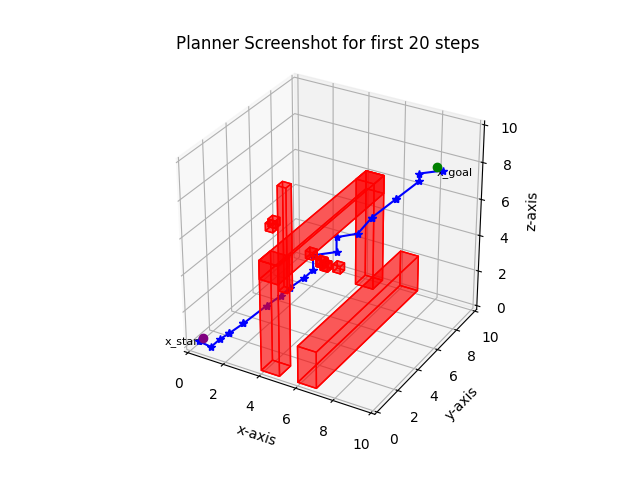}
    \includegraphics[width=7.5cm]{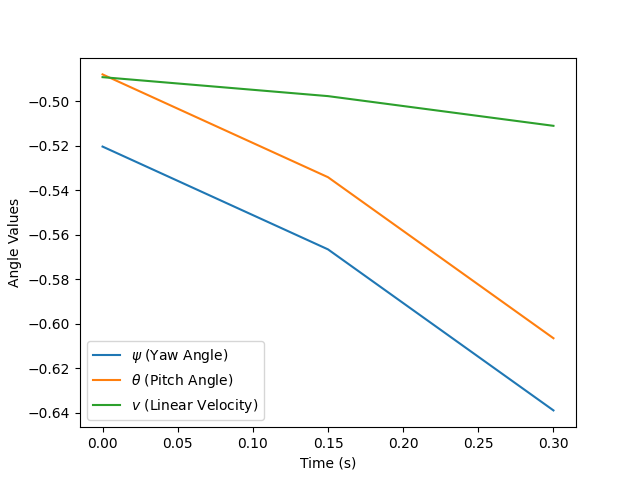}
    \caption{An illustration of the Planner, involving kinodynamics constraints \eqref{drone}, on the dynamic 3D environment evolving according to Case $4$ as the agent progresses step by step until reaching $x_{goal}$. The motion progression is displayed on the left, with the corresponding control inputs applied shown on the right.}
    \label{efig11}
\end{figure}

Similarly, the results of the Planner test in cases $1$ and $4$, as considered in the 3D example, are shown in Figs. \ref{efig8}, \ref{efig9}, \ref{efig10}, and \ref{efig11}. In these figures, the \texttt{`zorder'} functionality in the Python module \texttt{`matplotlib'} has been utilized to visualize the occlusion of the path that passes behind or in front of an obstacle bar. The workspace has been extended as $X=[0,10]^3$. However, in this example, we have explored situations where the planner has to enforce the agent's kinodynamic constraint, and the drone dynamics presented in \eqref{drone} has been implemented in this setting.

\begin{equation}
    \label{drone}
    \begin{cases}
        \dot{x} = v \cos(\psi) \cos(\theta) \\
        \dot{y} = v \sin(\psi) \cos(\theta) \\
        \dot{z} = v \sin(\theta) \\
        \dot{\psi} = \omega \\
        \dot{\theta} = \alpha \\
        \dot{v} = a,
    \end{cases}
\end{equation}
where $(x, y, z)\in X$ is the position coordinates in the workspace, $\psi\in[-\frac{\pi}{6},\frac{\pi}{6}]$ is the Yaw angle, representing the rotation about the vertical axis, $\theta\in[-\frac{\pi}{6},\frac{\pi}{6}]$ is the Pitch angle, representing the rotation about the lateral axis,
$v\in[-\frac{1}{2},\frac{1}{2}]$ is the linear velocity magnitude,
$\omega\in[-\frac{\pi}{2},\frac{\pi}{2}]$ is the angular velocity about the vertical axis, $\alpha\in[-\frac{\pi}{2},\frac{\pi}{2}]$ is the angular acceleration about the lateral axis, and $a\in[-1,1]$ is the linear acceleration magnitude. The Runge-Kutta $4$ method of the \texttt{`scipy'} functionality in Python has been adopted to solve \eqref{drone} over a time span of $[0s,0.3s]$ to simulate the drone dynamics for the subsequent state.

Additionally, similar to \eqref{3dyn}, we adopted the dynamics presented in \eqref{3dyn} to represent the involved adversarial dynamic obstacles employed in Case $4$ for the 3D example. 
\begin{equation}
    \label{3dyn}
    \begin{cases}
        x(k+1)=x(k)+\mathcal{R}([-h,h])\\
        y(k+1)=y(k)+\mathcal{R}([-h,h])\\
        z(k+1)=z(k)+\mathcal{R}([-h,h]).
    \end{cases}
\end{equation}

From the figures, it is evident that there might have been a proposed path for the agent, but due to the dynamic evolution of the environment, the planner eventually adjusts the initial path to prevent the agent from colliding with newly emerging obstacles. Remarkably, since randomness is involved in the way the environment evolves, similar but not necessarily identical figures will be generated when compiling the Python scripts attached to this report.

The completeness and optimality of the planner in this dynamic environment context depend on the nature of the adversarial modifications introduced to the environment and the specific characteristics of the traditional planner incorporated into the algorithm. Completeness, in this context, refers to the planner's ability to find a solution, if one exists, within a finite amount of time. The planner exhibits completeness by actively seeking solutions within the distortion time, striving to identify a path during this period, potentially overcoming adversarial obstacles, and progressing toward the goal. However, the adversarial nature of the obstacle may result in complete blockage, preventing the planner from reaching the goal and stopping mid-way, especially given the compact nature of the workspace.

Furthermore, the optimality of the planner is contingent on the nature of the adversarial dynamics that govern the modification of the environment. If the adversarial player introduces obstacles in a manner that strategically hinders the planner, there may be scenarios where no optimal path exists within the given distortion time. The optimization of the path is further influenced by the choice of the traditional planner involved.

\section{Conclusion}
This work is specifically designed to address the challenges posed by dynamic and constantly changing environments. The proposed algorithm exhibits exceptional navigation capabilities within familiar and confined areas, skillfully handling sudden appearance of obstacles, and excelling in situations involving adversarial obstacle dynamics. Its effectiveness is highlighted across a range of applications, from urban navigation to security-related operations, emphasizing its ability to adaptively respond to environmental shifts and adversarial challenges.

In particular, tackling dynamic environments in planning requires integrating a temporal dimension into the state space, a strategy employed in this work that incorporates time steps after adjusting a planned motion. However, this approach introduces challenges, including considerably longer planning times because of the expanded number of states that must be explored. The continuous and periodic changes in the environment require rapid plan generation, as delays could render the plans obsolete before they can be executed. In practical applications, this approach may sometimes lead to suboptimal outcomes, as the algorithm might choose longer paths around dynamic obstacles, overlooking more efficient options, such as waiting for the obstacle to move.

Moreover, in certain scenarios, the proposed approach might not be able to guide the agent to its intended destination. For example, if the trajectory of a dynamic obstacle intersects or blocks a doorway essential for the robot to reach its goal, as seen in cases $3$ and $4$ of the evaluated scenarios, the navigation may fail. Despite these limitations, this work provides valuable insights into navigating dynamic environments and establishes a foundation for further refinement and research in this critical area. Future work could examine incorporating additional specifications, such as linear temporal logic or broader $\omega$-regular properties \cite{baier2008principles} (or a subset thereof), which involve finite visits to specific regions of the state space \cite{ajeleye2024bdata}, and extending these properties to environments with stochastic evolution \cite{ajeleye2024co}.

\section{Acknowledgment}
The author extends gratitude to Morteza Lahijanian for insightful discussions and guidance throughout this work, as well as for the class sessions that inspired it.

\bibliographystyle{alpha}
\bibliography{biblio.bib}

\newcommand{\etalchar}[1]{$^{#1}$}
\begin{thebibliography}{XWD{\etalchar{+}}12}

\bibitem[AJE20]{ajeleye2020time}
DANIEL~AJEDAMOLA AJELEYE.
\newblock Time optimal output feedback control of nondeterministic finite state machines with safe reachability.
\newblock 2020.

\bibitem[ALZ23]{ajeleye2023data}
Daniel Ajeleye, Abolfazl Lavaei, and Majid Zamani.
\newblock Data-driven controller synthesis via finite abstractions with formal guarantees.
\newblock {\em IEEE Control Systems Letters}, 7:3453--3458, 2023.

\bibitem[AMP22]{ajeleye2022output}
Daniel~Ajedamola Ajeleye, Tommaso Masciulli, and Giordano Pola.
\newblock Output feedback control of nondeterministic finite--state systems with reach--avoid specifications.
\newblock In {\em 2022 30th Mediterranean Conference on Control and Automation (MED)}, pages 1012--1017. IEEE, 2022.

\bibitem[AZ24a]{ajeleye2024co}
Daniel Ajeleye and Majid Zamani.
\newblock Co-b{\"u}chi control barrier certificates for stochastic control systems.
\newblock {\em IEEE Control Systems Letters}, 2024.

\bibitem[AZ24b]{ajeleye2024cdata}
Daniel Ajeleye and Majid Zamani.
\newblock Data-driven construction of finite abstractions for interconnected systems: A compositional approach.
\newblock {\em arXiv preprint arXiv:2408.08497}, 2024.

\bibitem[AZ24c]{ajeleye2024bdata}
Daniel Ajeleye and Majid Zamani.
\newblock Data-driven controller synthesis via co-b{\"u}chi barrier certificates with formal guarantees.
\newblock {\em IEEE Control Systems Letters}, 2024.

\bibitem[BIP05]{belta2005discrete}
Calin Belta, Volkan Isler, and George~J Pappas.
\newblock Discrete abstractions for robot motion planning and control in polygonal environments.
\newblock {\em IEEE Transactions on Robotics}, 21(5):864--874, 2005.

\bibitem[BK08]{baier2008principles}
Christel Baier and Joost-Pieter Katoen.
\newblock {\em Principles of model checking}.
\newblock MIT press, 2008.

\bibitem[C{\etalchar{+}}47]{cauchy1847methode}
Augustin Cauchy et~al.
\newblock M{\'e}thode g{\'e}n{\'e}rale pour la r{\'e}solution des systemes d’{\'e}quations simultan{\'e}es.
\newblock {\em Comp. Rend. Sci. Paris}, 25(1847):536--538, 1847.

\bibitem[Dij22]{dijkstra2022note}
Edsger~W Dijkstra.
\newblock A note on two problems in connexion with graphs.
\newblock In {\em Edsger Wybe Dijkstra: his life, work, and legacy}, pages 287--290. 2022.

\bibitem[HNR68]{hart1968formal}
Peter~E Hart, Nils~J Nilsson, and Bertram Raphael.
\newblock A formal basis for the heuristic determination of minimum cost paths.
\newblock {\em IEEE transactions on Systems Science and Cybernetics}, 4(2):100--107, 1968.

\bibitem[KF11]{karaman2011sampling}
Sertac Karaman and Emilio Frazzoli.
\newblock Sampling-based algorithms for optimal motion planning.
\newblock {\em The international journal of robotics research}, 30(7):846--894, 2011.

\bibitem[Kha86]{khatib1986real}
Oussama Khatib.
\newblock Real-time obstacle avoidance for manipulators and mobile robots.
\newblock {\em The international journal of robotics research}, 5(1):90--98, 1986.

\bibitem[KL00]{kuffner2000rrt}
James~J Kuffner and Steven~M LaValle.
\newblock Rrt-connect: An efficient approach to single-query path planning.
\newblock In {\em Proceedings 2000 ICRA. Millennium Conference. IEEE International Conference on Robotics and Automation. Symposia Proceedings (Cat. No. 00CH37065)}, volume~2, pages 995--1001. IEEE, 2000.

\bibitem[KSLO96]{kavraki1996probabilistic}
Lydia~E Kavraki, Petr Svestka, J-C Latombe, and Mark~H Overmars.
\newblock Probabilistic roadmaps for path planning in high-dimensional configuration spaces.
\newblock {\em IEEE transactions on Robotics and Automation}, 12(4):566--580, 1996.

\bibitem[LaV98]{lavalle1998rapidly}
Steven LaValle.
\newblock Rapidly-exploring random trees: A new tool for path planning.
\newblock {\em Research Report 9811}, 1998.

\bibitem[LKJ01]{lavalle2001randomized}
Steven~M LaValle and James~J Kuffner~Jr.
\newblock Randomized kinodynamic planning.
\newblock {\em The international journal of robotics research}, 20(5):378--400, 2001.

\bibitem[PF05]{petti2005safe}
St{\'e}phane Petti and Thierry Fraichard.
\newblock Safe motion planning in dynamic environments.
\newblock In {\em 2005 IEEE/RSJ International Conference on Intelligent Robots and Systems}, pages 2210--2215. IEEE, 2005.

\bibitem[PHC{\etalchar{+}}21]{pairet2021online}
{\`E}ric Pairet, Juan~David Hern{\'a}ndez, Marc Carreras, Yvan Petillot, and Morteza Lahijanian.
\newblock Online mapping and motion planning under uncertainty for safe navigation in unknown environments.
\newblock {\em IEEE Transactions on Automation Science and Engineering}, 19(4):3356--3378, 2021.

\bibitem[SZ22]{salamati2022safety}
Ali Salamati and Majid Zamani.
\newblock Safety verification of stochastic systems: A repetitive scenario approach.
\newblock {\em IEEE Control Systems Letters}, 7:448--453, 2022.

\bibitem[XWD{\etalchar{+}}12]{xu2012real}
Wenda Xu, Junqing Wei, John~M Dolan, Huijing Zhao, and Hongbin Zha.
\newblock A real-time motion planner with trajectory optimization for autonomous vehicles.
\newblock In {\em 2012 IEEE International Conference on Robotics and Automation}, pages 2061--2067. IEEE, 2012.

\end{thebibliography}
\end{document}